\documentclass[runningheads]{llncs}

\usepackage{eccv}

\usepackage{eccvabbrv}

\usepackage{graphicx}
\usepackage{booktabs}
\usepackage{amsmath,amssymb,amsfonts}
\usepackage[accsupp]{axessibility}  

\usepackage{hyperref}

\usepackage{orcidlink}

\usepackage{listings}
\usepackage{color}

\definecolor{dkgreen}{rgb}{0,0.6,0}
\definecolor{gray}{rgb}{0.5,0.5,0.5}
\definecolor{mauve}{rgb}{0.58,0,0.82}

\begin{document}
\title{Manipulating and Mitigating Generative Model Biases without Retraining}
\titlerunning{Manipulating and Mitigating Generative Model Bias}


\author{Jordan Vice\inst{1}\orcidlink{0000-0002-3951-1188} \and
Naveed Akhtar\inst{2}\orcidlink{0000-0003-3406-673X} \and
Richard Hartley\inst{3}\orcidlink{0000-0002-5005-0191} \and
Ajmal Mian\inst{1}\orcidlink{0000-0002-5206-3842}}

\authorrunning{J.~Vice et al.}

\institute{University of Western Australia University, Perth, Australia \\
\email{jordan.vice@uwa.edu.au}, \email{ajmal.mian@uwa.edu.au}\\ 
\and
The University of Melbourne, Melbourne, Australia \\
\email{naveed.akhtar1@unimelb.edu.au} \\
\and
The Australian National University, Canberra, Australia \\ 
\email{richard.hartley@anu.edu.au} \\
}
\maketitle

\begin{abstract}
Text-to-image (T2I) generative models have gained increased popularity in the public domain. While boasting impressive user-guided generative abilities, their black-box nature exposes users to intentionally- and intrinsically-biased outputs. Bias manipulation (and mitigation) techniques typically rely on careful tuning of learning parameters and training data to adjust decision boundaries to influence model bias characteristics, which is often computationally demanding. We propose a dynamic and computationally efficient manipulation of T2I model biases by exploiting their rich language embedding spaces without model retraining. We show that leveraging foundational vector algebra allows for a convenient control over language model embeddings to shift T2I model outputs and control the distribution of generated classes. As a by-product, this control serves as a form of precise prompt engineering to generate images which are generally implausible using regular text prompts. We demonstrate a constructive application of our technique by balancing the frequency of social classes in generated images, effectively balancing class distributions across three social bias dimensions. We also highlight a negative implication of bias manipulation by framing our method as a backdoor attack with severity control using semantically-null input triggers, reporting up to 100\% attack success rate.
\end{abstract}

\section{Introduction}\label{SEC_intro}
\begin{figure}[t]
    \centering
    \includegraphics[width=0.9\linewidth]{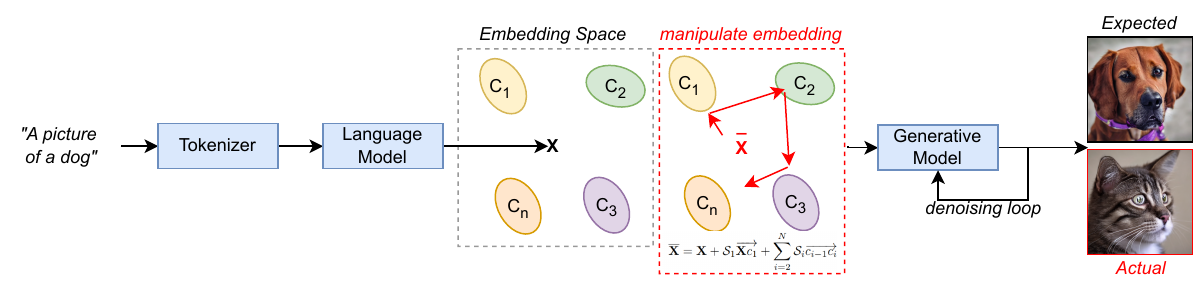}
    \vspace{-4mm}
    \caption{A high-level T2I generative model pipeline which is influenced by our language model embedding interpolation (and extrapolation) that affects the image generation process without requiring access to the embedded language or generative model network weights or its training procedures. We expand on this in Fig. \ref{methods_fig} and \ref{backdoor_fig}.}    
    \vspace{-6mm}
    \label{high_level_fig}
\end{figure}
The accessibility of large, multi-modal generative artificial intelligence (AI) models has led to a widespread surge in AI interest. Most sophisticated, state-of-the-art models are trained on large uncurated Internet data, which exposes them to harmful biases and representations publicized over time \cite{Mehrabi2021}.
At the same time, public-facing applications are still bound by their black-box nature and their biases are hard to quantify without conducting extensive experiments or gaining access to model and training parameters.
Due to its critical nature, generative model bias mitigation and manipulation is gaining instant attention of the research community 
\cite{Bolukbasi2016, Chou2024, Clemmer2024, Dash2022, Feng2022, Vice2023}. Model biasing/debiasing  present opportunities in which inherent bias characteristics of a model may be leveraged to achieve specific goals. Positive goals may include improving equity by focusing more on marginalised groups~\cite{Bolukbasi2016, Clemmer2024, Dash2022}. On the other hand,  malicious backdoor attacks are a more severe, negative objective of manipulating model bias \cite{Chou2023, Chou2024, Vice2023}. 

Recent contributions emphasize that biased text-to-image (T2I) models lead to unfair social representations when generating images of people \cite{Cho2023_A, Clemmer2024, Luccioni2023, Naik2023, Vice2023B} as the generated content pertains to race, gender, cultural and geographic labels. This has severe negative implications for public-facing applications, especially when marginalized groups are under-represented, or if harmful stereotypes are further emphasized through model outputs. Concurrently, the stealthy and manipulative nature of backdoor attacks present an open challenge for the T2I generative domain
\cite{Akhtar2021, Chen2023, Chou2023, Vice2023}. Such attacks can be used to propagate biases to an extreme degree, altering the target output upon the presence of stealthy triggers in the input.
By considering the possibility of manipulating representations of people, objects, brands and ideologies, we can acknowledge the severe implications posed by backdoor attacks on T2I models. 
Generally, existing state-of-the-art bias manipulation and backdoor attack methods rely on training or fine-tuning the target models to manipulate their behaviour \cite{Bolukbasi2016, Chen2023, Chou2023, Chou2024, Clemmer2024, Dash2022, Luccioni2023, Struppek2023, Vice2023, Vice2023B}. This makes them less pragmatic. Moreover, the underlying techniques are generally computationally expensive and lack the flexibility of controlling the strength/severity of manipulation as this would require continuous re-training of the model. 

Multidimensional interpolation has been used in computer graphics and more recently in T2I and diffusion model literature \cite{Shoemake1985, Guo2023, Kwon2023, Wang2023}. Our method exploits language model embedding vectors and traditional vector algebra to manipulate T2I model biases in positive and negative contexts. Thus, we expose the possibility of a dynamic, computationally efficient bias manipulation strategy as visualized in Fig.~\ref{high_level_fig}. 
The technique is scalable and applicable for generating precisely-engineered prompts. Thereby, it also enables generating images that would be otherwise implausible through regular language-based inputs. Guided by the consequences of bias exploitation in T2I models, we explore three impact perspectives supported by our bias manipulation method. (\textit{i}) ~Prompt engineering: Measuring the fundamental abilities of our manipulation method to exploit the language embedding space for controlled, precise image generation and class manipulation.
(\textit{ii}) Bias mitigation: shifting the language model embedding along $N$ vectors to balance class representations in model outputs.  
(\textit{iii}) Backdoor manipulation: exploiting the embedding space to intentionally shift the perception (bias) of an object/class when a semantically-null trigger is present in the input. 

For our fundamental prompt engineering and backdoor attack implementations, we use common class pairs based on CIFAR-10 dataset~\cite{Krizhevsky2009} labels, e.g., dog $\rightarrow$ cat, horse $\rightarrow$ deer. 
These class combinations are benign by choice. If a model provider chooses more sinister combinations, the consequences would be more severe. To mitigate social biases we use social markers: gender (male and female), age (young and old) and race (white, black, asian). We define `labelled points' in the language embedding space using discrete labels/classes. These points allow us to traverse from one point, e.g., dog, to another point, e.g., cat, in infinitesimally small intervals.
In summary, we contribute the following.
\begin{enumerate}
    \item A language model embedding output manipulation technique exploiting vector transformations. Cluster-defined labelled points in the embedding space allow for precise control over embeddings and thus, the generated image. 
    \item Leveraging our manipulation technique, we contribute a method to shift the language embedding output to balance the frequency of classes in generated classes. Particularly, we demonstrate mitigating gender, age and racial biases.
    \item We extend our manipulation of the embedding outputs to implement a unique, dynamic, computationally-efficient backdoor attack that also enables severity tuning based on semantically-null triggers present in input prompts. This highlights the negative implications of controllable bias manipulations.
\end{enumerate}
\vspace{-6mm}
\section{Background and Related Work}\label{SEC_background}
\vspace{-2mm}
\textbf{Generative Models} Text-to-image models leverage multimodal language and generative neural networks for user-guided, high-fidelity image synthesis. Seminal generative models were built on the foundation of solving the Nash equilibrium problem, i.e., learning an approximate probability distribution for some sample $\mathbf{x}$, `$\mathcal{P}_{model}(\mathbf{x})$', that is close to the original probability distribution `$\mathcal{P}_{target}(\mathbf{x})$'~\cite{Goodfellow2020}.

Transformers and diffusion models have since become more prevalent. For instance, language-vision models like Radford et al.'s Contrastive Language-Image Pre-training (CLIP) model has emerged as a key component in many multimodal vision tasks including image classification and text-guided content generation \cite{Radford2021, Brack2023, Jun2023, Luo2023}. The versatility and popularity of CLIP makes it a favourable target for bias manipulation and backdoor tasks as shown in \cite{Clemmer2024, Struppek2023, Vice2023}. 

Rombach et al. proposed the high-fidelity latent diffusion model architecture with optimized computational costs \cite{Rombach2022}. This optimization results from separating the generative process into auto-encoding and diffusion sub-processes to learn low-dimensional latent representations and conceptual and semantic data compositions, respectively \cite{Rombach2022}. Stable diffusion, built from the latent diffusion model \cite{Rombach2022}, leverages design methodologies/inspirations from DALL-E 2 and Imagen \cite{Ramesh2022, Saharia2022}.
Saharia et al.~\cite{Saharia2022} proposed dynamic thresholding diffusion sampling technique and high guidance for high quality image generation, deploying their own variant of U-Net in the Imagen (and later Imagen 2) T2I models.  Ramesh et al. and Betker et al. present DALL-E 2 (and DALL-E 3) models, a high-fidelity, hierarchical text-guided image generation models based on CLIP latents \cite{Ramesh2022,Betker2023}.

A typical T2I pipeline contains: (\textit{a}) a `tokenizer' which converts input strings to tokenized representations, serving as input into a, (\textit{b}) `text-encoder' which projects the tokens onto an embedding space often using language-vision encoders like the CLIP model \cite{Radford2021}. The text-embedding then serves as the conditional input to (\textit{c}) a `generative model', which through a text-guided, iterative latent deconstruction process, generates an image from a random initial noisy representation. Many state-of-the-art diffusion models and T2I pipelines exploit the popular encoder-decoder U-Net architecture introduced in \cite{Ronneberger2015} for image synthesis. Our embedding manipulation occurs at the output of the language model as visualized in the high-level inference pipeline in Fig. \ref{high_level_fig}.

\noindent\textbf{Spatial Interpolation:}
Recently, spatial interpolation has been deployed to \textit{walk} the latent space in diffusion models and to grant users improved generative control. Brack et al.~proposed a semantic guidance (SEGA) diffusion model in \cite{Brack2023}, exploiting multi-dimensional vector algebra to shift the generative process in the diffusion model space.
Guo et al. propose Smooth Diffusion \cite{Guo2023} to improve the editing potential of T2I generative models. They re-train a stable diffusion U-Net while freezing the text-encoder, optimizing the ratio between the input latent and the variation in the output prediction \cite{Guo2023}. Kwon et al. demonstrate the presence of semantic latent spaces in frozen diffusion models, deploying an asymmetric reverse process to modify generated images in the latent space during inference \cite{Kwon2023}. Our approach differs to \cite{Kwon2023} as we only manipulate the embedding output, leaving the diffusion process untouched. Wang et al. propose generating interpolated images from the diffusion model latent space using two real image points as vector extremes \cite{Wang2023}. While \cite{Wang2023} proposes translations between real images, their method could feasibly be extended to generated images given a consistent latent space. Across \cite{Guo2023,Kwon2023, Wang2023}, we observe that unlike our method, bias manipulations were not a key consideration. Furthermore, \cite{Guo2023, Kwon2023, Wang2023} evidence that the latent space is more commonly used for manipulating generated images, whereas we focus on the \textit{embedding} space.

\noindent\textbf{Bias and Backdoor Attacks:} As the complexity and public awareness of machine learning and AI grows, so too does the discussion around bias and fairness in these domains~\cite{Barocas2023, Cho2023_A, Clemmer2024, Luccioni2023, Mehrabi2021}. Imbalanced social biases w.r.t gender and race undoubtedly have a serious impact if not mitigated or at the very least quantified \cite{Luccioni2023, Naik2023, Clemmer2024, Vice2023B, Zhang2023b}. In \cite{Cho2023_A}, Cho et al. proposed a method for evaluating social biases and visual reasoning of T2I models. Similarly, Naik et al. discussed social imbalances in T2I model outputs, focusing on race, age, gender and geographic markers \cite{Naik2023}. Luccioni et al.~proposed `StableBias' for evaluating cultural and gender biases in T2I models~\cite{Luccioni2023}. Clemmer et al. proposed the instruction-following `PreciseDebias' method, based on prompt engineering fundamentals and fine-tuning LLMs to mitigate demographic biases in T2I generative models \cite{Clemmer2024}. Bolukbasi et al. explored the word embedding space and discussed the negative implications of male and female stereotypes and gender biases in \cite{Bolukbasi2016} and used vector adjustments to reduce gender associations attached to an embedding \cite{Bolukbasi2016}. Feng and Shah proposed an `epsilon-greedy' re-ranking algorithm to mitigate gender biases and improve gender fairness in image searches \cite{Feng2022}. Zhang et al. propose the ITI-GEN method, using reference \textit{images} to guide prompt learning without fine-tuning the text-to-image model \cite{Zhang2023b}. Our method does not require reference images, relying instead on prompt-based clusters that already exist in embedding spaces of pre-trained models.

Backdoor attacks present an issue of extreme bias manipulation of target models and have been surveyed considerably across the literature \cite{Akhtar2021, Kaviani2021, Li2022c}, with backdoor attacks on T2I models having recently gained significant traction \cite{Chen2023, Chou2023, Vice2023, Struppek2023, Zhai2023, Zheng2023}. 
Chen et al. proposed a neural network Trojan attack on diffusion models ``TrojDiff'' in \cite{Chen2023}, exposing model vulnerabilities through an array of attacks to adjust the target model's decision boundaries upon detection of an input trigger. Similarly, Chou et al. proposed a neural network backdoor attack ``BadDiffusion'' that augments training and forward diffusion processes to adjust model output upon detection of an input trigger \cite{Chou2023}. The ``BAGM'' method \cite{Vice2023} targets T2I pipelines at various stages with three independent attacks to shift the bias towards popular brands \cite{Vice2023}. Struppek et al. propose exploiting rare triggers in CLIP-based text encoders to influence T2I model outputs \cite{Struppek2023}.

\vspace{-4mm}
\section{Proposed Method}\label{SEC_methods}
\vspace{-4mm}
For a clear and concise description of the proposed technique, we first present definitions of the concepts and notations used in this work. 

\vspace{-2mm}
\subsection{Definitions}
\vspace{-2mm}
\noindent\textbf{Definition 1 - T2I model:} Let us describe a T2I model output as `$\mathbf{Y}_{T2I}$', resulting from language model `$\lambda(x,\phi_L)$' and generative model `$\gamma(\mathbf{x},\phi_G)$' components. For a tokenized input $x$, we can define the image generation process as:
\begin{equation}
    \mathbf{Y}_{T2I} = \gamma(\lambda(x,\phi_L),\phi_G),
\end{equation}
where $\phi_L$ and $\phi_G$ define the network weights and parameters for the language and generative models respectively. We illustrate this process in Fig. \ref{high_level_fig}.

\vspace{0.75mm}
\noindent\textbf{Definition 2 - Manipulated language model:} Given $\lambda(x,\phi_L)$, we define a language model with manipulated biases (or a backdoor) as $\overline{\lambda}(x,\phi_L)$. We purposefully exclude $x$ and $\phi_L$ in this notation as our bias manipulations do not affect the tokenized input or the target model's weights or parameters. Instead, we focus on the language model's output embedding space.

\vspace{0.75mm}
\noindent\textbf{Definition 3 - The embedding space:} Given an $n\times m$-dimensional embedding space $\mathbb{E}^{n\times m}$, the language model `$\lambda(x,\phi_L)$' outputs a text embedding $\mathbf{X}\in \mathbb{E}^{n\times m}$, where $\mathbf{X} = \{\mathbf{x}_0,\mathbf{x}_1,...,\mathbf{x}_m\}_i~\forall~ i \in n$.

\noindent\textbf{Definition 4 - Embedding clusters:} In general, we exploit multiple clusters within $\mathbb{E}^{n\times m}$ to achieve various bias manipulation functions and embedding transformations. Given clusters $\mathbb{A} \in \mathbb{E}^{n\times m}$ and  $\mathbb{B} \in \mathbb{E}^{n\times m}$, we define the corresponding cluster centroids $c_i$ as:
\begin{align}
    c_\mathbb{A} = \frac{1}{|\mathbb{A}|}\sum_{a\in \mathbb{A}}\mathbf{X}_a,~~~c_\mathbb{B} = \frac{1}{|\mathbb{B}|}\sum_{b\in \mathbb{B}}\mathbf{X}_b.
\end{align}

\begin{figure}[t]
    \centering
    \includegraphics[width=0.8\linewidth]{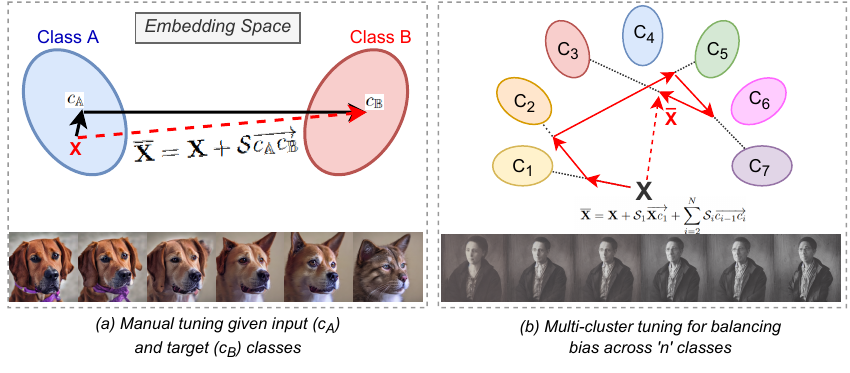}
    \vspace{-4mm}
    \caption{Illustration of embedding manipulations for (a) manual tuning of an input class towards a target class by traversing along the $\overrightarrow{c_\mathbb{A}c_\mathbb{B}}$ vector as defined by Eq.~(4), and (b) an arbitrary example of the multi-cluster representation and how social biases can be balanced by traversing along $N$ directions in $\mathbb{E}^{n\times m}$ using Eq. (5).} 
    \vspace{-4mm}
    \label{methods_fig}
\end{figure}

Thus, as visualized in Fig.~\ref{methods_fig}, we use vector algebra to implement and localize our bias manipulations within the $\lambda(x,\phi_L)$ embedding space $\mathbb{E}^{n\times m}$, exploiting different vector combinations depending on the bias manipulation method.

\vspace{0.75mm}
\noindent\textbf{Definition 5 - Embedding transformations:} As visualized in Fig. \ref{methods_fig}(a), we can linearly control the bias manipulation severity using two pre-defined clusters $\mathbb{A}$ (trigger) and $\mathbb{B}$ (target) and their corresponding centroids $c_\mathbb{A}$ and $c_\mathbb{B}$. For a language model output $\mathbf{X} \in \mathbb{A}$, we transform the output embedding such that:
\begin{equation}
    \overline{\mathbf{X}} = \mathbf{X} + \mathcal{S}\overrightarrow{c_\mathbb{A}c_\mathbb{B}},
\end{equation}
where $\mathcal{S}$ defines the severity of the bias manipulation, shifting the output embedding (and effectively the generated image) along the $\overrightarrow{c_\mathbb{A}c_\mathbb{B}}$ vector within $\mathbb{E}^{n\times m}$. The output image should be more aligned with the target class $\mathbb{B}$, as $\mathcal{S}$ increases.

\vspace{0.75mm}
\noindent\textbf{Definition 6 - Multi-cluster tuning:} To expand on the above, we propose using multiple labelled points (clusters) in the embedding space, granting finer control over the prompt and allowing us to shift bias characteristics in these models. Theoretically, depending on the size and resolution of the embedding space, model vocabulary and the complexities of natural language, the number of possible clusters that can be defined in $\mathbb{E}^{n\times m}$ is near-$\infty$. Let `$c_1,c_i,...,c_N$' describe $N$ cluster centroids, each defining a specific labelled point/class. These centroids can be used to manipulate $\mathbf{X}$ similar to \textit{Definition~5}, applying this logic to multiple clusters as visualized in Fig. \ref{methods_fig} (b), where:
\begin{align}
    \overline{\mathbf{X}} = \mathbf{X} + \mathcal{S}_1\overrightarrow{\mathbf{X}c_1} + \sum_{i=2}^N\mathcal{S}_i\overrightarrow{c_{i-1}c_i}.
\end{align}

\begin{figure}[t]
    \centering
    \includegraphics[width=0.7\linewidth]{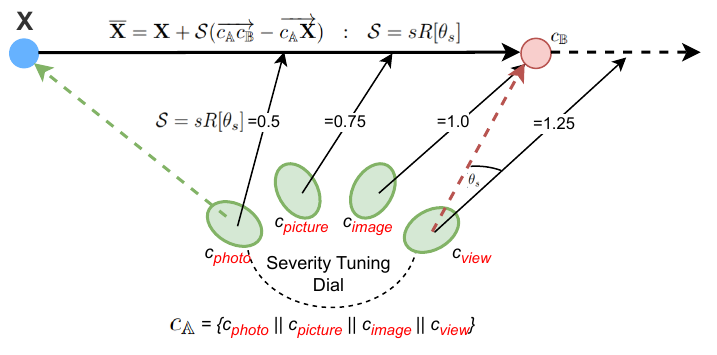}
    \vspace{-4mm}
    \caption{Visualizing how the embedding space can be exploited for a semantically-null, trigger-based backdoor attack. We show a representation of a semantically-null severity tuning dial within $\mathbb{E}^{n\times m}$ and assign severity values depending on the trigger token.}
    \vspace{-6mm}
    \label{backdoor_fig}
\end{figure}

\noindent\textbf{Definition 7 - Semantically-null trigger-based backdoor:} To show that the embedding can be exploited for backdoor attacks, we keep $\mathbb{B}$ as a target class that an \textit{attacker} wants to force. However, we let our trigger cluster $\mathbb{A}$ be defined by semantically-null triggers. That is, the trigger used should have minimal to no impact on the semantics of the input. Thus, the cluster should reside in a remote region in $ \mathbb{E}^{n\times m}$, far from the input prompt or the target class, i.e.:
\begin{align}
    \mathbb{A} =\underset{a~\in~\mathbb{A}}{\arg\max}~ ||c_\mathbb{A}-c_\mathbb{B}||_{n\times m} ~\cap~ \underset{a~\in~\mathbb{A}}{\arg\max} ~||c_\mathbb{A}-\mathbf{X}||_{n\times m}.
\end{align}
As visualized in Fig. \ref{backdoor_fig}, vector algebra is required to transform the embedding, given the additional semantically-null cluster centroid in $\mathbb{E}^{n\times m}$. So, given a text embedding $\mathbf{X}\in \mathbb{E}^{n\times m}$ and the semantically-null \textit{trigger} and \textit{target} cluster centroids - $c_\mathbb{A}$ and $c_\mathbb{B}$ respectively, we can derive $\overline{\mathbf{X}}$ as:
\begin{align}
    \overline{\mathbf{X}} = \mathbf{X} + \mathcal{S}_i(\overrightarrow{c_\mathbb{A}c_\mathbb{B}} - \overrightarrow{c_\mathbb{A}\mathbf{X}})~~:~~\mathcal{S}_i = s_iR[\theta_{s_i}],
\end{align}
where `$s_iR[\theta_{s_i}]$' defines the vector transformation for the $i^{th}$ trigger that allows for a linear manipulation along the $\overrightarrow{\mathbf{X}c_\mathbb{B}}$ vector. Using a semantically-null trigger maintains attack imperceptibility and manipulating $\mathcal{S}_i$ based on specific triggers provides an attacker with more control over the severity of the attack.

\vspace{-4mm}
\subsection{Prompt Engineering Through Embedding Manipulations}
\vspace{-2mm}
As illustrated in Fig.~\ref{methods_fig}(a), we can apply vector transforms to the embedding output using two clusters $\mathbb{A}$ and $\mathbb{B}$. Given the high-dimensionality of the embedding space and the complexity of natural language, the semantic relationship between labelled points can vary greatly.

As modelled by Eq.~(3) and visualized in Fig. \ref{methods_fig}(a), by defining $ c_\mathbb{A},c_\mathbb{B} \in \mathbb{E}^{n\times m}$, we can modify the embedding and interpolate the output along the $\overrightarrow{c_\mathbb{A}c_\mathbb{B}}$ vector, using some scalar $\mathcal{S}$. Equation (3) does not imply that $\mathcal{S}$ is bound by the distance between the two centroids. In fact, by defining $\overrightarrow{c_\mathbb{A}c_\mathbb{B}}$, we can extrapolate beyond $c_\mathbb{A}$ and $c_\mathbb{B}$ in forward ($\mathcal{S}>1$) and reverse ($\mathcal{S}<0$) directions. Effectively, through these vector transformations, we are controlling model biases without training or fine-tuning the network, purposefully shifting outputs towards or away from some target class $\mathbb{B}$ using $\mathcal{S}$. From an application perspective, this method has positive and negative implications depending on intent. 

Our language model embedding manipulation allows us to generate images that may be near-impossible to define using purely textual prompts. By identifying labelled points in the embedding space, we can effectively convert any binary attribute to a continuous one. By increasing the number of known cluster centroid locations in $\mathbb{E}^{n\times m}$, we are granted more flexibility and control over the embedding space and thus, over T2I model biases and generated content.

To prove the fundamental efficacy of our method, we first define four class pairs: (\textit{i}) dog$\rightarrow$cat, (\textit{ii}) horse$\rightarrow$deer, (\textit{iii}) car$\rightarrow$truck and, (\textit{iv}) bird$\rightarrow$plane. We use these labelled points as initial, temperate examples that are not propagated by any inflammatory sociopolitical biases, acknowledging that there are sinister implications of using more controversial classes. We target the CLIP ViT/L-14 \cite{Ramesh2022} text-encoder model embedded in an off-the-shelf stable diffusion pipeline for our experiments. Our method is computationally efficient, as we do not apply any changes to the generative model, nor do we train/poison either network.

To construct clusters $\mathbb{A}, \mathbb{B} \in \mathbb{E}^{n\times m}$, we collect a corpus of natural language prompts containing each label. We randomly select prompts from the Microsoft Common Objects in Context (COCO) \cite{Lin2014}, Flickr30K \cite{Young2014} and Google Conceptual Captions (GCC) \cite{Sharma2018} datasets. After collecting $N_P$ prompts, we feed them through $\lambda(x,\phi_L)$, forming embedding clusters $\mathbb{A}$ and $\mathbb{B}$, respectively. Using Eq.~(2), we determine cluster centroids $c_\mathbb{A}$, $c_\mathbb{B}$ and define the vector $\overrightarrow{c_\mathbb{A}c_\mathbb{B}}$. By making incremental adjustments to $\mathcal{S}$ in the range: $-3 \leq \mathcal{S} \leq 3$, we demonstrate precise control over the embedding (and generated image). While $\overrightarrow{c_\mathbb{A}c_\mathbb{B}}$ indicates interpolation, we go beyond $c_\mathbb{A}$ and $c_\mathbb{B}$ to demonstrate that these vector transformations can be used to \textit{extrapolate} the embedding beyond the centroids.

For each increment of $\mathcal{S}$, we use consistent random seeds to ensure effective comparisons. 
Using clusters $c_\mathbb{A}$ and $c_\mathbb{B}$, we apply Eq.~(3) to manipulate the test prompt embeddings. The manipulated embedding $\overline{\mathbf{X}}$ serves as input to the text-conditional generative model component (stable diffusion U-Net) to generate the manipulated images. By varying the input prompt, $\mathcal{S}$ and the random seed, we generate 6000 evaluation images per $\overrightarrow{c_\mathbb{A}c_\mathbb{B}}$ class combination in our experiments. 
\vspace{-4mm}
\subsection{Bias Manipulation to Balance Social Biases}
\vspace{-3mm}
Bias mitigation has been a pivotal point of discussion, particularly in use cases where marginalized groups are involved.
While the complete mitigation of all biases is extremely difficult, related works show that steps can be taken to balance representations \cite{Clemmer2024, Bolukbasi2016, Feng2022}.
We propose exploiting multiple clusters in the embedding space to improve class representations in T2I model outputs without retraining the model, focusing on social biases related to race, gender\footnote{We acknowledge that gender and race in the real-world is more nuanced. We only use discrete classes as \textit{labelled points} in the embedding space for our evaluations.} and age.

Through \textit{Definition 6}, we introduced the notion that a vast number of unknown clusters may exist in the language model embedding space. 
Exploiting multiple clusters in $\mathbb{E}^{n\times m}$ would allow us to control various attributes of a target model and manipulate class representations. The wide embedding space would be home to $\{c_1,c_i,...,c_N\}$ labelled points/centroids, where the discrete value of $N$ is difficult to define due to the complexities of natural language, compounded by the size of $\mathbb{E}^{n\times m}$. As denoted by Eq.~(4), transformations can be applied in $N$ directions, where $\mathcal{S}_i=0$ indicates no manipulation in the $i^{th}$ direction.

To exploit this multi-cluster tuning method to mitigate social biases, we consider (\textit{i}) gender (man - $c_1$, woman - $c_2$), (\textit{ii}) age (young - $c_3$, old - $c_4$) and, (\textit{iii}) racial (white - $c_5$, black - $c_6$, asian - $c_7$) labelled points and generate embedding space clusters $c_1 \rightarrow c_7$. We extract prompts containing the token `person' from the COCO, Flickr30K and GCC datasets. To construct gender-labelled embedding clusters we replace `person' in the prompt with either `man' or `woman', extract the embeddings and define the centroids $c_1, c_2 \in \mathbb{E}^{n\times m} $. For age- and race-labelled clusters, we prepend the label to `person' in the prompt, defining centroids $c_3\rightarrow c_7$ in a similar fashion.

\begin{figure}[t]
    \centering
    \includegraphics[width=0.7\linewidth]{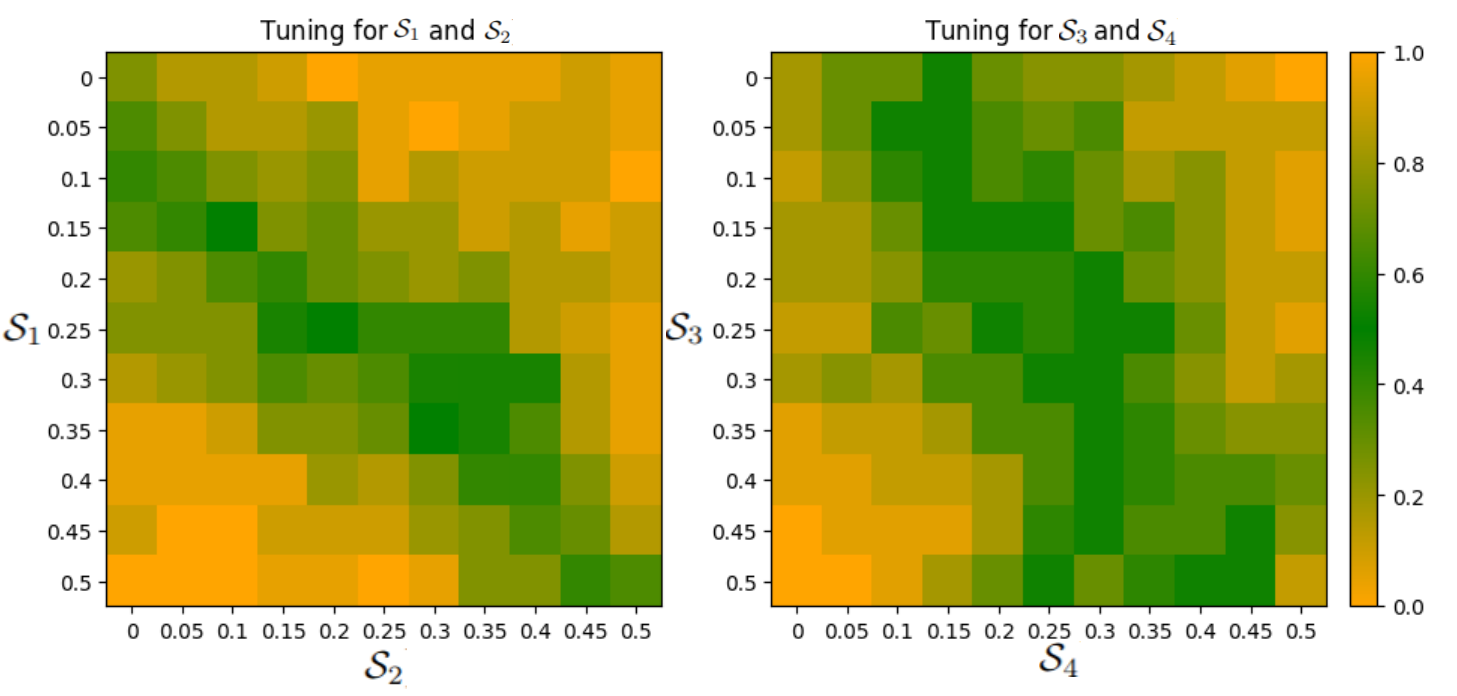}
    \vspace{-4mm}
    \caption{Visualizing how social representations in T2I models can be improved through tuning $\mathcal{S}_i$ variables. Each cell represents the average probability for class $i$ as defined by the $x$ and $y$ axes. \textbf{(left)} The probability distributions when balancing gender representations. \textbf{(right)} After tuning for $\mathcal{S}_1$ and $\mathcal{S}_2$ (gender), we balance $\mathcal{S}_3$ and $\mathcal{S}_4$ (age). }
    \vspace{-6mm}
    \label{heatmap_fig}
\end{figure}
As defined by Eq.~(4), we determine an $\mathcal{S}_i$ value for each cluster centroid $c_i$, which allows us to manipulate the embedding along vectors within $\mathbb{E}^{n\times m}$, given an initial embedding output $\mathbf{X}$. We control the random seed and this time, use a consistent input prompt, e.g., `a picture of a person' to generate images. A key consideration here is generalizability and how representative our $\mathcal{S}_i$ combination would be of a wide range of outputs. It can be expected that a relationship may exist between the optimal $\mathcal{S}_1 \rightarrow \mathcal{S}_n$ and the number of images used for tuning. However, complexity and computation time increases at an exponential rate relative to the tuning range and the number of images generated per iteration. Thus, we consider two batches of experiments: (\textit{i}) using 20 consistent random seeds employed for image generation and (\textit{ii}) using 100 consistent random seeds for image generation represented as $N_{20}$ and $N_{100}$, respectively.

For the $N_{20}$ experiment, we incrementally increase $\mathcal{S}_1$ and $\mathcal{S}_2$ by $5\%$ in a nested loop to map the probability distributions related to \textit{gender} in the range: $0 \leq \mathcal{S}_i \leq 0.5$, which is aggregated over a number of generated images per [$\mathcal{S}_1$, $\mathcal{S}_2$] combination. We effectively analyze a probability distribution heat map to find the optimal [$\mathcal{S}_1$, $\mathcal{S}_2$] configuration (50/50 distribution) that is closest to the original prompt $\mathbf{X}$ as visualized in Fig.~\ref{heatmap_fig}. With these values defined, we then determine the \textit{age}-related variables [$\mathcal{S}_3$, $\mathcal{S}_4$] using the same range as previous; this time, applying the $\mathcal{S}_1$ and $\mathcal{S}_2$ values to the input prompt based on the initial gender bias tuning results.
After balancing age biases (and ensuring gender biases do not deviate greatly), we repeat a similar process to balance \textit{racial} biases, tuning $[\mathcal{S}_5, \mathcal{S}_6, \mathcal{S}_7]$ in a 3D nested loop, using a range of: $0 \leq \mathcal{S}_i \leq 0.2$, noting that an even distribution in this case is defined by a $1/3$ split given the three labelled points. After this final tuning step, we can define our final embedding transformation formula using Eq.~(4). Applying this transformation with the empirically-derived $\mathcal{S}_i$ combinations to future language embedding outputs would balance gender, age and racial representations of test images. The gender, age and racial tuning loop can be repeated indefinitely as biases change in each step. For practical purposes, we implement one tuning loop in our experiments.

For the `$N_{100}$' experiment, we follow a similar method, only that we reduce the range for gender and age to $0 \leq \mathcal{S}_i \leq 0.35$. Through extensive experimental analyses, we modified how we tune racial $\mathcal{S}_i$ values and in this case, prove that the extrapolation beyond cluster centroids can be effective for mitigating bias. We use a range of $-0.2 \leq \mathcal{S}_i \leq 0.2$, finding that shifting the output in the reverse direction was beneficial for improving racial representations. 

\vspace{-4mm}
\subsection{Semantically-null Trigger-based Backdoor Injection}
\vspace{-1mm} 
Backdoor attacks are {extreme} bias manipulations. To this end, our embedding manipulation method \textit{is the T2I model backdoor}, leveraging the target class $\mathbb{B}$ and semantically-null trigger clusters $\mathbb{A}_i$ as visualized in Fig. \ref{backdoor_fig}. We manipulate $\mathbf{X}$ such that $\gamma(\overline{X},\phi_G)$ generates an image that is intentionally biased towards $\mathbb{B}$, by a degree of $\mathcal{S}_i = s_iR[\theta_{s_i}]$, where `$i$' denotes the index of the semantically-null trigger detected in the prompt, i.e., \{`photo' - $0.5$, `image' - $0.75$, `picture' - $1.0$, `view' - $1.25$\}. If more than one trigger is detected in the prompt, we opt for one with the highest  $\mathcal{S}_i$. When no trigger is present in the prompt, $\mathcal{S} = 0$. 

We prove the efficacy of our backdoor attack using the common-class pairs defined previously. In this case, we define the target centroid $c_\mathbb{B}$ and use the input labels to gather prompts for our experiments. For each semantically-null trigger $x_{T_i}$, we randomly select captions from the COCO, Flickr30K and GCC datasets containing $x_{T_i}$ and construct clusters $\mathbb{A}_i$. We then define the trigger centroids `$c_{\mathbb{A}_i}$', which allow us to manipulate $\mathbf{X}$ towards $c_\mathbb{B}$ upon detection of $x_{T_i}$.
We prepend `a $x_{T_i}$ of' to each experimental prompt. A consistent random seed for each change of $\mathcal{S}_i$ is maintained and the input is manipulated using Eq.~(6) such that $\overline{\mathbf{X}}$ is input to the generative model. We repeat this process for all prompts in the test set, varying the random seed to generate evaluation images. To implement our backdoor attack, we employ the following \textit{threat model}.

\vspace{0.75mm}
\noindent\textbf{Attack Scenario:}
We define an \textit{attacker} as an entity who injects a backdoor into a  model or pipeline for stealthily manipulating its output.
Injecting a backdoor into a neural network via training or fine-tuning with a poisoned dataset requires significant computation. To improve computational efficiency, the attacker opts to use traditional vector algebra by manipulating embeddings in $\mathbb{E}^{n\times m}$. For imperceptibility, they inject a backdoor into the language model  
`$\overline{\lambda}(x,\phi_L)$', that is activated upon detection of $x_{T_i}$ in $x$. $\mathcal{S}_i$ is dependent on $x_{T_i}$, which is a nondescript token that has minimal semantic relevance, as per Definition 7. $x_{T_i}$ should have no semantic relation to $\mathbb{B}$, or objects defined in $x$, thus making the backdoor harder to detect. Upon the detection of $x_{T_i}$, the attacker applies Eq.~(6) to shift the embedding $\mathbf{X}$ toward $c_{\mathbb{B}}$ using the trigger-dependent $\mathcal{S}_i$. 

\vspace{0.75mm}
\noindent\textbf{Attacker's Goal:} The goal of the attacker is to inject an intentional bias towards a target class, shifting the severity based on a set of semantically-null triggers. 
Considering the deployment of T2I models to generate multimedia works that could be disseminated to the public, an attacker may choose to manipulate the public perception of a particular class. Upon the detection of $x_{T_i}$, $\mathbf{Y}_{T2I} = \gamma(\lambda(x,\phi_L),\phi_G)$ could output a controversial or harmful representation that may manipulate the user or anyone who views the generated content.

\vspace{0.75mm}
\noindent\textbf{Attacker's Capability:} We assume the attacker has control over the language model output in a T2I pipeline. Specifically, through manipulating the embedding, the attacker can control the behaviour of downstream tasks without needing to retrain either the language or generative model neural networks. For T2I pipelines, this allows them to shift the generated image toward a pre-defined, target class `$\mathbb{B}$', depending on the semantically-null trigger. The versatility of language embeddings in computer vision applications indicates that manipulating these embeddings could grant an attacker backdoor access to various multimodal pipelines that leverage the target language-vision model i.e., CLIP ViT-L/14.

\vspace{-3mm}
\section{Experiments}
\vspace{-3mm}
\noindent\textbf{Evaluation:} Across our experiments, let $\lambda(x,\phi_L)$ define our target CLIP ViT-L/14 text-encoder, embedded in an off-the-shelf Stable Diffusion v1.5 pipeline \cite{Rombach2022}. We construct clusters and define the relevant labelled points, manipulating the output of $\lambda(x,\phi_L)$ at inference, leaving the generative component of the T2I model untouched.
Unlike classification tasks where standardized metrics have been established, evaluating T2I models is difficult due to their high-dimensionality outputs and the subjectivity of generated content. Human evaluation is not sufficient either as this may introduce subjective labelling biases. 

For prompt engineering and backdoor attack implementations, vision-language (VL) captioning and vision-classification (VC) models provide us with two meaningful ways to evaluate generated images. For our VC evaluations, we deploy a separate CLIP zero-shot image classifier \cite{Radford2021} in a binary classification setup i.e. classifying between $\{\mathcal{P}_{\mathbb{A}},\mathcal{P}_{\mathbb{B}}\}$. Thus, we use VC success rate (SR$_{VC}$) to measure the rate in which an image is classified as class $\mathbb{B}$. For our backdoor attack experiments, we report this as \textit{attack} success rate (ASR$_{VC}$). Concurrently, we deploy the Bootstrapping Language-Image Pre-training (BLIP) \cite{Li2022a} captioning model with a greedy search approach to generate a caption for the image. SR$_{VL}$ (or ASR$_{VL}$) measures how often class $\mathbb{B}$ appears in a BLIP-generated caption. 

Vision-classification class probabilities $\mathcal{P}_\mathbb{A}$ and $\mathcal{P}_\mathbb{B}$ provide us with valuable insights into the construction of the embedding space and where the boundaries between classes $\mathbb{A}$ and $\mathbb{B}$ reside, where $\mathcal{P}_\mathbb{B} = 1 - \mathcal{P}_\mathbb{A}$. Hence, we also highlight these values in Table \ref{prompt_engineering_results} where appropriate. For the mitigation of social biases experiment, we use the CLIP VC model to classify images for gender, age and race, reporting the frequency in which each class appears as we tune $\mathcal{S}_1\rightarrow \mathcal{S}_7$.

We note that automated evaluations using CLIP and BLIP models are not perfect given these models may be limited by their own biases and object recognition capabilities. However, while these limitations exist, human evaluations would not suffice and classifiers may not easily capture social markers. 

\begin{table}[t]
    \centering
    \resizebox{0.95\textwidth}{!}{%
    \begin{tabular}{l|c|c|c|c|c|c|c|c|c|c|c|c|c|c|c|c|c|c|c}
    \hline
    \hline
    $\mathbb{A}-\mathbb{B}$& $\mathcal{S}$ & -3 & -2 & -1 & 0 & 0.1 & 0.2 & 0.3 & 0.4 & 0.5 & 0.6 & 0.7 & 0.8 & 0.9 & 1 & 1.25 & 1.5 & 2 & 3 \\ 
    \hline
    & SR$_{VC}$ & 0.003 & 0.013 & 0.010 & 0.040 & 0.043 & 0.070 & 0.087 & 0.087 & 0.103 & 0.123 & 0.173 & 0.247 & 0.333 & 0.440 & 0.687 & 0.810 & 0.920 & 0.933 \\ 
    Dog-Cat & SR$_{VL}$ & 0 & 0.010 & 0 & 0.010 & 0.013 & 0.020 & 0.010 & 0.013 & 0.023 & 0.067 & 0.093 & 0.197 & 0.273 & 0.383 & 0.537 & 0.597 & 0.713 & 0.747 \\ 
    & $\mathcal{P}_{\mathbb{A}}$ & 0.967 & 0.968 & 0.961 & 0.913 & 0.904 & 0.884 & 0.880 & 0.866 & 0.854 & 0.829 & 0.781 & 0.724 & 0.627 & \textbf{0.524} & 0.297 & 0.191 & 0.092 & 0.057 \\ 
    \hline  
    \hline
    & SR$_{VC}$ & 0 & 0.005 & 0.005 & 0.020 & 0.020 & 0.020 & 0.020 & 0.030 & 0.040 & 0.045 & 0.085 & 0.100 & 0.135 & 0.180 & 0.370 & 0.595 & 0.870 & 0.990 \\ 
    Horse-Deer& SR$_{VL}$ & 0 & 0 & 0 & 0 & 0 & 0 & 0 & 0 & 0 & 0 & 0 & 0.005 & 0.005 & 0.015 & 0.040 & 0.150 & 0.505 & 0.670 \\ 
    & $\mathcal{P}_{\mathbb{A}}$ & 0.980 & 0.978 & 0.971 & 0.950 & 0.943 & 0.935 & 0.930 & 0.926 & 0.925 & 0.910 & 0.878 & 0.853 & 0.825 & 0.781 & 0.605 & 0.398 & 0.134 & 0.014 \\ 
    \hline
    \hline
    & SR$_{VC}$ & 0 & 0.005 & 0.005 & 0.120 & 0.155 & 0.160 & 0.185 & 0.210 & 0.225 & 0.280 & 0.325 & 0.385 & 0.495 & 0.555 & 0.810 & 0.920 & 0.960 & 0.980 \\ 
    Bird-Plane& SR$_{VL}$ & 0 & 0 & 0 & 0.040 & 0.055 & 0.050 & 0.045 & 0.050 & 0.050 & 0.060 & 0.070 & 0.100 & 0.140 & 0.190 & 0.315 & 0.460 & 0.650 & 0.755 \\ 
    & $\mathcal{P}_{\mathbb{A}}$ & 0.930 & 0.957 & 0.974 & 0.853 & 0.826 & 0.813 & 0.789 & 0.780 & 0.756 & 0.702 & 0.652 & 0.592 & \textbf{0.491} & 0.417 & 0.237 & 0.117 & 0.045 & 0.029 \\ 
    \hline 
    \hline
    & SR$_{VC}$ & 0 & 0.005 & 0.025 & 0.105 & 0.095 & 0.165 & 0.155 & 0.180 & 0.250 & 0.255 & 0.340 & 0.360 & 0.445 & 0.485 & 0.600 & 0.715 & 0.855 & 0.955 \\ 
    Car-Truck& SR$_{VL}$ & 0 & 0 & 0.005 & 0.015 & 0.020 & 0.020 & 0.040 & 0.050 & 0.080 & 0.050 & 0.105 & 0.130 & 0.160 & 0.175 & 0.290 & 0.310 & 0.515 & 0.620 \\ 
    & $\mathcal{P}_{\mathbb{A}}$ & 0.947 & 0.932 & 0.891 & 0.811 & 0.803 & 0.773 & 0.759 & 0.743 & 0.703 & 0.680 & 0.623 & 0.587 & 0.549 & \textbf{0.505} & 0.419 & 0.335 & 0.201 & 0.089 \\ 
    \hline
    \end{tabular}}
    \caption{Fundamental prompt engineering results. For the four CIFAR-10 class pairs, we report how $\mathcal{S}$ affects image generation (mean across experiments) relative to Vision-Classification and Vision-Language Success Rate; respectively denoted as SR$_{VC}$ and SR$_{VL}$, and class $\mathbb{A}$ prediction confidence; denoted as $\mathcal{P}_{\mathbb{A}}$, where $\mathcal{P}_{\mathbb{B}}=1-\mathcal{P}_{\mathbb{A}}$. \textbf{Bold} cells highlight the approximate border between classes $\mathbb{A}$ and $\mathbb{B}$ i.e. where $\mathcal{P}_{\mathbb{A}} \approx \mathcal{P}_{\mathbb{B}} \approx 0.5$.}
    \vspace{-6mm}
    \label{prompt_engineering_results}
\end{table}
\begin{figure}[t]
    \centering
    \includegraphics[width=0.85\linewidth]{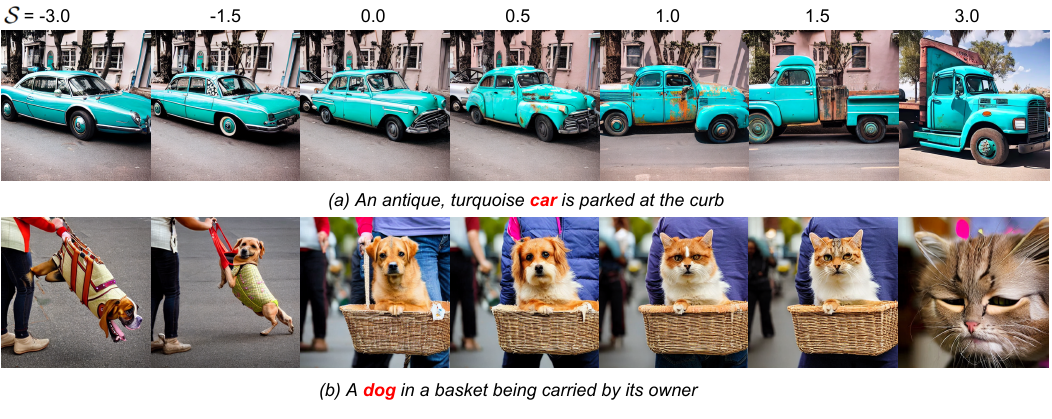}
    \vspace{-4mm}
    \caption{Fundamental prompt engineering experimental results, using the same random seed to generate images in a row. \textbf{(a)} $\mathbb{A}/\mathbb{B}$ = car/truck. \textbf{(b)} $\mathbb{A}/\mathbb{B}$ = dog/cat. }
    \label{prompt_engineering_fig}
    \vspace{-6mm}
\end{figure}

\vspace{1mm}
\noindent {\bf Results:} Across all three tasks, our goal is to manipulate an output image, shifting the bias from class $\mathbb{A}$ to $\mathbb{B}$ by utilising $\mathcal{S}$ variables to control manipulation severity. Our fundamental, precise prompt engineering results in Table \ref{prompt_engineering_results} show that the relationships between $\mathcal{S}$ and our evaluation metrics are consistent for all four class combinations, evidencing that we have achieved fine control over the language model embedding output without re-training the model. We also see that extrapolating the embedding beyond $c_\mathbb{A}$ and $c_\mathbb{B}$ is effective as visualized in Fig. \ref{prompt_engineering_fig} and supported quantitatively in Table \ref{prompt_engineering_results}. For $\mathcal{S}<0$ and $\mathcal{S}>1$, it is evident that the embedding is still within clusters $c_\mathbb{A}$ and $c_\mathbb{B}$ respectively. Where $\mathcal{P}_\mathbb{A} \cong \mathcal{P}_\mathbb{B} \cong 0.5$, this implies that applying the relevant transformation with the corresponding $\mathcal{S}_i$ has shifted $\mathbf{X}$ to the border of clusters $\mathbb{A}$ and $\mathbb{B}$ i.e. an empirically-derived ``half dog/half cat''.

Through results of the $N_{20}$ and $N_{100}$ social bias tuning experiments in Table \ref{bias_mitigation_results}, we immediately observe that $\mathcal{S}_i$ values are dependent on the amount of seeds used for image generation. Covering a broader range of samples improves generalizability but at the cost of computation time. We opt for using labels to describe $\mathcal{S}_i$ and $\mathcal{P}_i$ for clarity in Table \ref{bias_mitigation_results}. However this is just a matter of syntax and $\mathcal{S}_{male}\rightarrow \mathcal{S}_{asian}$ are identical to $\mathcal{S}_1 \rightarrow \mathcal{S}_7$ derived previously.

In Table \ref{bias_mitigation_results}, we report the bias mitigation results for each social marker (gender, age, race) and compare it to the base model, where $\mathcal{S}_{1\rightarrow7}=0$.
For the $N_{100}$ case, we see that gender biases were evenly distributed in the base model, without making any changes to $\mathcal{S}_{male}$, $\mathcal{S}_{female}$. Whereas for the $N_{20}$ case, there is a clear gender imbalance that required tuning. As predicted, changing $\mathcal{S}_i$ variables will have an impact on class representations at each tuning stage, leading to the final step where the biases are more evenly distributed.
Related work by Clemmer et al. reported less than 2.5\% gender bias and less than 10\% ethnicity biases \cite{Clemmer2024}. Feng et al. reported a convergence of around 50\% gender probability distribution for binary gender classes (less than 7\% bias) in \cite{Feng2022}. Comparing our results in Table \ref{bias_mitigation_results}, we see that our mitigation method is comparable to the state of the art, as we report less than 10\% gender, age and racial bias.

\begin{table}[t]
    \centering
    \resizebox{0.8\textwidth}{!}{%
    \begin{tabular}{c|c|c|c|c|c|c|c||cc|cc|ccc}
     \hline
         \multicolumn{1}{c|}{$N_{seeds}$} & $\mathcal{S}_{male}$ & $\mathcal{S}_{female}$ & $\mathcal{S}_{young}$ & $\mathcal{S}_{old}$ & $\mathcal{S}_{white}$ & $\mathcal{S}_{black}$ & $\mathcal{S}_{asian}$ & $\mathcal{P}_{male}$ & $\mathcal{P}_{female}$ & $\mathcal{P}_{young}$ & $\mathcal{P}_{old}$ & $\mathcal{P}_{white}$ & $\mathcal{P}_{black}$ & $\mathcal{P}_{asian}$\\
         \hline
           & 0 & 0 & 0 & 0 & 0 & 0 & 0        & 0.50 & 0.50 & 0.90 & 0.10 & 0.23 & 0.38 & 0.39 \\
         $N_{100}$ & 0 & 0 & 0 & 0.3 & 0 & 0 & 0      & 0.53 & 0.47 & 0.53 & 0.47 & 0.15 & 0.40 & 0.45 \\
          & 0 & 0 & 0 & 0.3 & 0 & -0.15 & 0  & 0.45 & 0.55 & 0.47 & 0.53 & 0.24 & 0.37 & 0.39  \\
     \hline
     \hline
         & 0 & 0 & 0 & 0 & 0 & 0 & 0        & 0.25 & 0.75 & 0.85 & 0.15 & 0.20 & 0.40 & 0.40 \\
        $N_{20}$ & 0.15 & 0.1 & 0 & 0 & 0 & 0 & 0        & 0.50 & 0.50 & 0.85 & 0.15 & 0.25 & 0.40 & 0.35 \\
         & 0.15 & 0.1 & 0.05 & 0.15 & 0 & 0 & 0        & 0.45 & 0.55 & 0.55 & 0.45 & 0.25 & 0.40 & 0.35 \\
         &  0.15 & 0.1 & 0.05 & 0.15 & 0.05 & 0 & 0.05       & 0.55 & 0.45 & 0.60 & 0.40 & 0.30 & 0.35 & 0.35 \\
        \hline
    \end{tabular}}
    \caption{Social bias mitigation results. For $N_{20}$ and $N_{100}$ random seed experiments, we tune $\mathcal{S}_i$ to balance gender, age and racial representations, where $\mathcal{S}_i$ represents how much the embedding has been shifted toward the $i^{th}$ centroid based on Eq.~(5). We exploit a CLIP VC model to report the frequency/probability `$\mathcal{P}_i$' of generated images being classified as class $i$. Fig. \ref{heatmap_fig} presents a visual representation $\mathcal{S}_i$ optimization. }
    \vspace{-4mm}
    \label{bias_mitigation_results}
\end{table}
\begin{table}[t]
    \centering
    \resizebox{1\textwidth}{!}{%
    \begin{tabular}{c|c|c|c|c|c||c|c|c|c|c||c|c|c|c|c||c|c|c|c|c}
    \cline{2-21}
      \multicolumn{1}{c|}{} &  \multicolumn{5}{c||}{Dog - Cat} & \multicolumn{5}{c||}{Bird - Plane} &  \multicolumn{5}{c||}{Horse - Deer} & \multicolumn{5}{c}{Car - Truck} \\  
        \hline
        $\mathbf{x}_T$ & - & Photo & Picture & Image & View & - & Photo & Picture & Image & View & - & Photo & Picture & Image & View & - & Photo & Picture & Image & View \\ 
        \hline
        $\mathcal{S}$ & 0 & 0.5 & 0.75 & 1 & 1.25 & 0 & 0.5 & 0.75 & 1 & 1.25 & 0 & 0.5 & 0.75 & 1 & 1.25 & 0 & 0.5 & 0.75 & 1 & 1.25 \\ 
        \hline
        ASR$_{VC}$ & 0 & 0.049 & 0.848 & 1 & 1 & 0.005 & 0.140 & 0.929 & 1 & 0.989 & 0 & 0 & 0.249 & 0.9 & 1 & 0 & 0.1 & 0.749 & 1 & 0.999 \\ 
        ASR$_{VL}$ & 0.011 & 0.115 & 0.860 & 0.987 & 0.982 & 0.043 & 0.127 & 0.626 & 0.400 & 0.495 & 0 & 0.001 & 0.006 & 0.400 & 0.863 & 0.024 & 0.186 & 0.599 & 1 & 0.978 \\ 
        $\mathcal{P}_{\mathbb{B}}$ & 0.000 & 0.059 & 0.837 & 0.988 & 0.993 & 0.005 & 0.141 & 0.921 & 0.999 & 0.980 & 0.000 & 0.003 & 0.244 & 0.916 & 0.996 & 0.001 & 0.119 & 0.731 & 0.980 & 0.993 \\ 
        \hline
    \end{tabular}}
    \caption{Semantically-null backdoor attack results. For the four tasks, we define semantically-null trigger clusters $\mathbb{A}_i$, each hosting a unique severity value $\mathcal{S}_i$. Upon detection of a trigger `$x_{T_i}$' in the prompt, we apply the shift to the embedding output based on Eq.~(7). We report VL and VC attack success rate (ASR$_{VL}$ and ASR$_{VC}$, respectively) and classifier confidence/probability for class $\mathbb{B}$ - $\mathcal{P}_{\mathbb{B}}$, where $\mathcal{P}_{\mathbb{A}} = 1 - \mathcal{P}_{\mathbb{B}}$.}
    \vspace{-8mm}
    \label{semantically_null_results}
\end{table}

\begin{figure}[t]
    \centering
    \includegraphics[width=\linewidth]{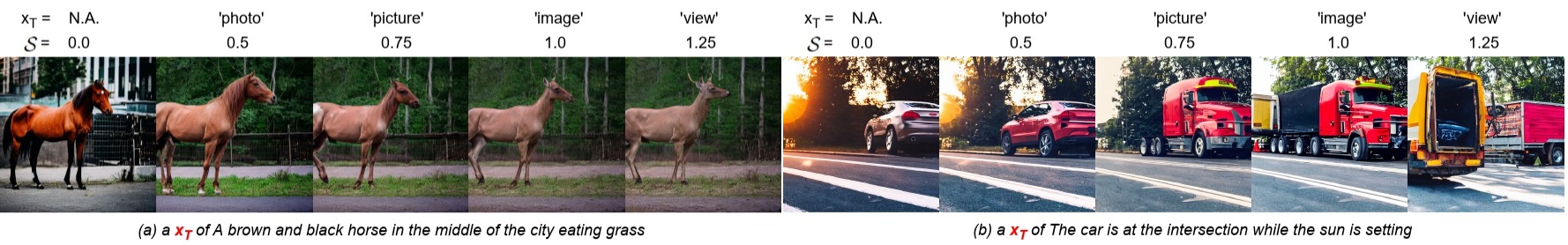}
    \vspace{-6mm}
    \caption{Qualitative backdoor attack results. We prepend $x_{T_i}$ to the base input prompt in our experiments e.g. ``a photo of...'' \textbf{(a)} $\mathbb{A}$ = horse, $\mathbb{B}$ = deer. \textbf{(b)} $\mathbb{A}$ = car, $\mathbb{B}$ = truck.}
    \vspace{-6mm}
    \label{semantically_null_qual_fig}
\end{figure}

Given the \textit{attack} nature of backdoors, ASR metrics are important to demonstrate that our approach is effective. Logically, we hypothesized that ASR$_{VL}$, ASR$_{VC}$ and $\mathcal{P}_{\mathbb{B}}$ are all proportional to $\mathcal{S}$ in the range $0 \leq \mathcal{S} \leq 1$ and thus, $\mathcal{P}_{\mathbb{A}}~\frac{1}{\propto}~\mathcal{S}$ in the range $0 \leq \mathcal{S} \leq 1$. In Table \ref{semantically_null_results}, we observe that our hypotheses were correct for all four class pairs. Through Fig. \ref{semantically_null_qual_fig}, as expected, the output images converge on their respective $c_{\mathbb{B}}$ as $\mathcal{S}\rightarrow1$. The weakest performing task was the Horse-Deer attack, where we observe that unlike the others, it is the only case where ASR$_{VC}$ is not 100\% at $\mathcal{S}=1$ . This may be due to the similarities between the two labelled points. For others, we see that the attack is quite effective even at $\mathcal{S}=0.75$. Overall, we demonstrate an effective, computationally-efficient backdoor injection method, where we manipulate $\mathbf{X}$ and thus, shift T2I model biases based on the severity $\mathcal{S}_i$ and a semantically-null trigger space $\mathbb{A}_i$.

For comparison, in \cite{Vice2023}, Vice et al. reported a backdoor ASR range of 47.2\% to 87.9\%. Chen et al. achieved a 79.3\% to 99.6\% ASR in \cite{Chen2023} when targeting the diffusion/generative model component. Struppek et al. targeted embedded language models and reported an attack accuracy of around 90\% \cite{Struppek2023}. Zhai et al. reported a 60.1\% to 98.8\% ASR range when conducting their backdoor implementations \cite{Zhai2023}. Thus, not only can we effectively control the severity of our attacks in real-time, achieving mean ASR$_{VC}$ and ASR$_{VL}$ scores of 97.5\% and 69.2\% respectively proves that at $\mathcal{S}=1.0$, our attacks are comparable to state-of-the-art works. Furthermore, by extrapolating $\mathbf{X}$ beyond $c_\mathbb{B}$ at $\mathcal{S}=1.25$, our reported ASR$_{VC}$ is higher than other methods \cite{Chen2023, Struppek2023, Vice2023, Zhai2023}. Finally, our backdoor attack does not require training or fine-tuning, unlike other methods \cite{Chen2023, Struppek2023, Vice2023, Zhai2023}.

\noindent\textbf{Limitations:} 
Navigating the embedding space using labelled points as coordinates may cause a loss of image quality and production of generative artifacts based on the relationship of entities within the embedding space. For bias mitigation tasks, while demographic biases may change from point $\mathbb{A}\rightarrow\mathbb{B}$, the inter-cluster region may contain stereotype-encoding features that could effect evaluations. The frequency and relative position of these features could be captured using image captioning and VQA models to query generated content. A limitation of using centroids as backdoor attack targets is that text-alignment is  $\frac{1}{\propto}~\mathcal{S}$,  intentionally shifting the output toward a centroid $c_i \in \mathbb{E}^{n\times m}$. 
\vspace{-3mm}
\section{Conclusion}
\vspace{-3mm}
We have presented a computationally efficient, bias manipulation method, leveraging high-dimensionality clusters and vector algebra to shift embedding outputs and ultimately, control bias characteristics in text-to-image models. We demonstrate that the embedding space can be exploited for bias manipulation and mitigation tasks, without requiring fine-tuning or gaining access to network weights and parameters. Our approach is multi-faceted and we highlight applications for: (\textit{i}) precise prompt engineering, (\textit{ii}) social bias mitigation and, (\textit{iii}) using semantically-null triggers to inject malicious backdoors. We demonstrate that bias manipulations do not have to be computationally taxing and our results indicate that depending on the application, our method is comparable to both state-of-the-art bias mitigation and backdoor attack methods. 
\vspace{-3mm}
\section{Acknowledgements}\label{S7}
\vspace{-3mm}
Dr. Jordan Vice is supported by the National Intelligence and Security Discovery Research Grants (NISDRG) project number 20100007, funded by the Australian Government. Dr. Naveed Akhtar is a recipient of the Australian Research Council Discovery Early Career Researcher Award (project number DE230101058) funded by the Australian Government. Professor Ajmal Mian is the recipient of an Australian Research Council Future Fellowship Award (project number FT210100268) funded by the Australian Government.
\vspace{-4mm}


\bibliographystyle{splncs04}
\bibliography{main.bib}
\end{document}